%% file: preprint.tex
\documentclass{article}

 \usepackage[preprint]{neurips_2026}

\usepackage[utf8]{inputenc} 
\usepackage[T1]{fontenc}    
\usepackage{hyperref}       
\usepackage{url}            
\usepackage{booktabs}       
\usepackage{amsfonts}       
\usepackage{nicefrac}       
\usepackage{microtype}      
\usepackage{xcolor}         

\usepackage{algorithm}
\usepackage{algorithmic}
\usepackage{amssymb}
\usepackage{amsmath}
\usepackage{enumitem}
\usepackage{multirow}
\usepackage[table]{xcolor}
\usepackage{makecell}
\usepackage{tabularx}
\usepackage{array}
\usepackage{graphicx}
\usepackage{wrapfig}

\title{Buried in Textual Debt: Context Pruning with Visual Evidence Preservation for MLLM Agents}

\author{
Yuchen Huang\thanks{Equal contribution.} \quad
Sijia Li\footnotemark[1] \quad
Jun Zhang\thanks{Corresponding author.} \quad
Yi R. (May) Fung \\
Hong Kong University of Science and Technology \\
\texttt{\{yhuanggn,slifg\}@connect.ust.hk}\\
\texttt{\{eejzhang,yrfung\}@ust.hk}
}

\begin{document}

\maketitle

\begin{abstract}
Multimodal Large Language Models (MLLMs) are increasingly deployed as multi-step agents, where explicit reasoning supports task decomposition and tool coordination but also accumulates self-generated text. Over long trajectories, this text can dominate the context and suppress visual evidence, creating \emph{textual debt}. We observe that reasoning becomes redundant once task-relevant visual evidence is grounded, while stale hypotheses can misguide later inference when grounding remains uncertain. Pruning must therefore remove redundant text without discarding visual evidence. We propose SPARE, a Kullback-Leibler (KL)-guided framework for pruning accumulated reasoning in multimodal tool-use agents. SPARE uses a compact task-state summary as privileged diagnostic context. For each candidate segment, it replays the same model under the original and summary-conditioned contexts. Reverse-KL divergence from on-policy self-distillation (OPSD) then tests whether the summary sufficiently covers the segment without disrupting future reasoning. We further fine-tune the summarizer with supervised fine-tuning (SFT), enabling more compact summaries, broader coverage, and more aggressive pruning. Across multi-step visual tool-use benchmarks, SPARE achieves the highest average accuracy among pruning methods while removing 37.89-64.58\% of reasoning tokens. This favorable accuracy-context trade-off shows that reducing textual dominance restores reliance on visual evidence and mitigates over-conditioning on self-generated language. Code is available at \url{https://github.com/lukahhcm/spare}.
\end{abstract}

\section{Introduction}
\label{sec:Introduction}

Multimodal Large Language Models (MLLMs) are increasingly deployed as agents that reason, call tools, inspect intermediate observations, and revise their answers over time \citep{li2025benchmark}. Explicit reasoning is beneficial in this setting, as it helps agents decompose visual tasks, coordinate multi-step tool use, interpret returned observations, and maintain progress across turns \citep{ke2025explain}. Yet it also introduces a largely overlooked cost: each interaction step adds self-generated text to the working context. Over long trajectories, this accumulated text can dominate computation relative to the original image and newly returned observations, making the model increasingly conditioned on its own plans, descriptions, and assumptions rather than external visual evidence.

Not all reasoning history remains useful throughout a multimodal reasoning trajectory. When early reasoning correctly captures task-relevant visual evidence, the accumulated text may still provide a faithful abstraction of the image. However, when early reasoning fails to attend to the relevant visual regions, the model may elaborate on an incomplete or weakly grounded textual state. In this case, additional reasoning tokens become \emph{textual debt}: stale linguistic context that contributes little new evidence while competing with image tokens for attention and context budget. From this view, pruning reasoning tokens is not merely a way to reduce computation or sequence length. It can also function as a grounding mechanism by reducing over-conditioning on stale textual context and restoring the opportunity to re-attend to the visual input.


Existing approaches do not directly address this problem. Visual-token compression reduces redundancy in the \emph{visual} stream by dropping or merging image tokens \citep{chen2024image,yang2025visionzip,shang2025llava,zhang2024sparsevlm,tan2025tokencarve}. While effective for reducing computation, it targets a modality that is often already under-attended in deep layers, and further compression can weaken the relative contribution of image evidence \citep{takezoe2026learnpruner}. Concise-reasoning methods primarily shorten newly generated rationales, while generic summarization compresses history at a coarse level. Neither identifies which historical segments remain functionally necessary. In multi-step agents, the more pressing source of redundancy is the self-generated text that accumulates across rounds.

\begin{figure*}[t]
    \centering
    \includegraphics[width=\textwidth]{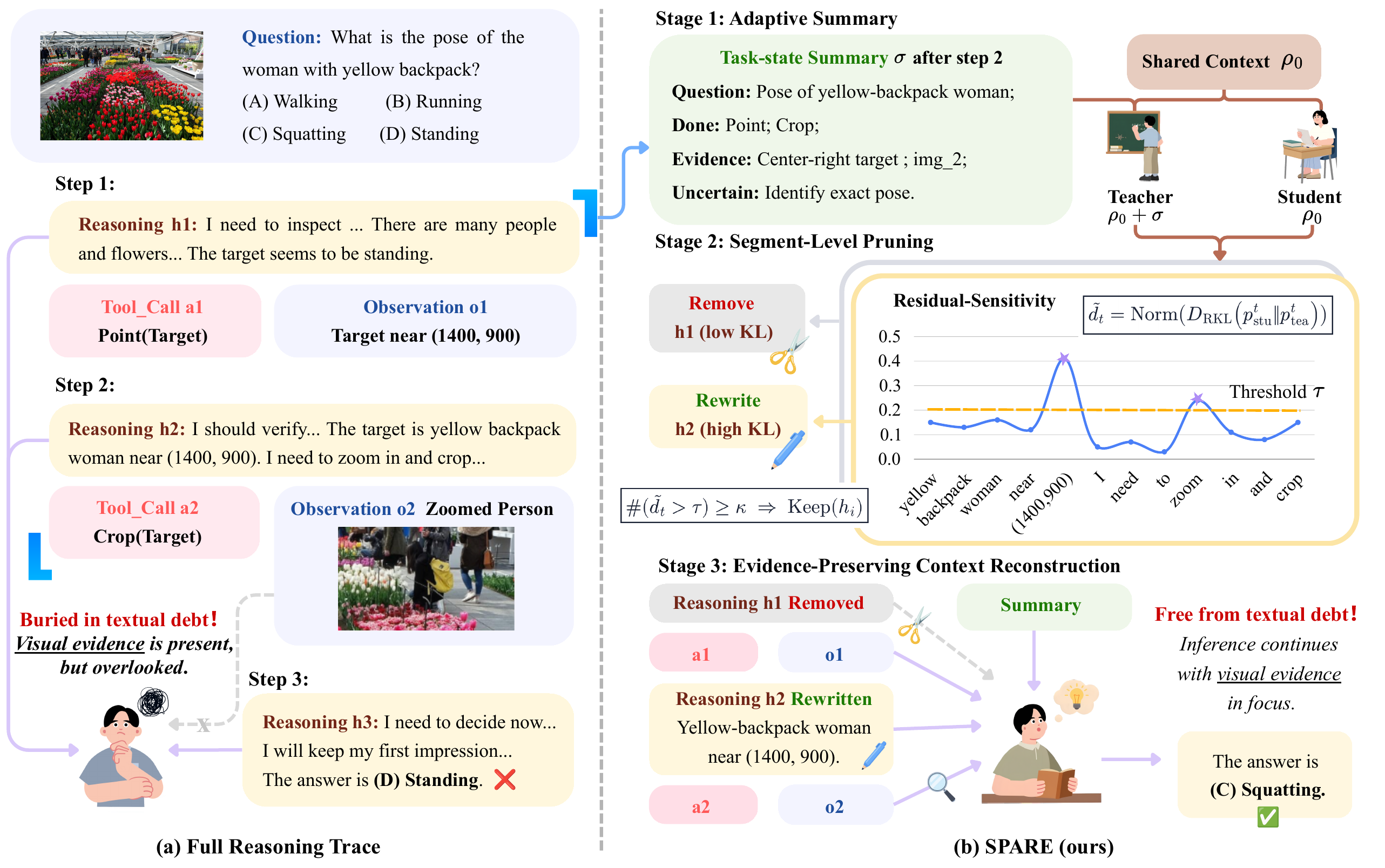}
\caption{
    Overview of SPARE. (a) In the full reasoning trace, accumulated self-generated text reinforces
    a stale hypothesis and obscures visual evidence returned by the tools,
    leading to an incorrect answer.
    (b) SPARE invokes a compact task-state summary as a \emph{transient probe}
    that is not written into the persistent trajectory. It replays each
    historical reasoning segment with and without summary conditioning and
    measures its token-level residual sensitivity using normalized reverse KL.
    A count-based rule prunes summary-covered segments with insufficient
    high-KL residue while retaining evidence-critical segments. Each pruned
    segment is replaced by concise structured visual evidence extracted from
    its original reasoning, whereas planning-dominated text is removed and the
    original tool calls, observations, and images remain unchanged. The
    resulting context reduces textual debt, refocuses subsequent inference on
    visual evidence, and produces the correct answer.
}
    \label{fig:spare_overview}
\end{figure*}

We therefore reframe context compression as a light-weight, \emph{evidence-preserving selective forgetting} problem and instantiate it as \textbf{SPARE} (Selective Pruning of Accumulated Reasoning with Visual Evidence Preservation), a KL-guided reasoning-pruning framework for multimodal tool-use agents. The key idea is to use a compact task-state summary as \emph{privileged diagnostic context} rather than as a replacement for the history: instead of overwriting the trajectory with the summary, we test whether each historical segment retains information beyond the consolidated state. Concretely, \textsc{SPARE} replays the same model under the original and summary-conditioned contexts and uses reverse-KL between token-level continuation distributions to measure residual dependence on the original segment. Segments with low residual sensitivity are pruned, while high-sensitivity segments are preserved as evidence-critical content. This diagnostic requires neither an external verifier nor a separate reward model. We further fine-tune the summarizer with supervised fine-tuning (SFT), enabling more compact summaries, broader coverage, and more aggressive pruning. Across multi-step visual tool-use benchmarks, \textsc{SPARE} achieves the highest average accuracy among pruning methods while removing 37.89-64.58\% of reasoning tokens. We also show that suppressing textual dominance increases attention to image tokens, supporting our claim that reducing textual debt restores reliance on visual evidence.

Our contributions are summarized as follows:
\begin{itemize}[leftmargin=*,itemsep=1pt,topsep=2pt]
    \item We propose \textsc{SPARE}, an evidence-preserving selective-forgetting method that uses summary-conditioned KL divergence to estimate the functional redundancy of prior reasoning and prune segments whose information has already been consolidated.
    \item We further fine-tune the summarizer with SFT to produce more compact task-state summaries, enabling broader coverage and more aggressive pruning while preserving task-relevant information.
    \item Experiments across multiple backbones and visual tool-use benchmarks show that \textsc{SPARE} improves the accuracy and restores the influence of visual evidence on downstream reasoning.
\end{itemize}

\section{Preliminaries}
\label{sec:Preliminaries}

\subsection{Multi-Step Tool Use in MLLMs}
\label{ssec:prelim_setup}

We consider a vision language model (VLM) $\pi_\theta$ that answers a user
question through multi-step, tool-augmented reasoning. The input consists of a
question $q$ and a set of image tokens $\mathcal{I}$, and the model may query a
fixed set of external tools $\mathcal{T}$ over at most $T$ interaction steps.

At step $t$, the model conditions on the current message history
$\mathbf{H}_t$ and produces an assistant message
\begin{equation}
    m_t = (h_t, a_t),
\end{equation}
where $h_t$ is a free-form reasoning span and $a_t$ is either a tool invocation
$\langle\text{tool\_call}\rangle$ or a final answer
$\langle\text{response}\rangle$. If $a_t$ invokes a tool, the corresponding
tool result is appended to the history, yielding $\mathbf{H}_{t+1}$, and the
interaction proceeds to the next step. If $a_t$ emits a final response, the
episode terminates. We denote the sequence of past reasoning-action segments
by $\mathcal{S}_t = \{\,s_i = (h_i, a_i)\,\}_{i<t}$.

\subsection{On-policy Self-Distillation}
\label{ssec:prelim_opsd}

Knowledge distillation trains a student model $\pi_\theta$ to match a teacher
distribution $\pi_{\mathrm{teacher}}$ over next tokens. Conventional
off-policy distillation applies this objective on fixed teacher or ground-truth
sequences, which creates exposure bias because the training prefixes differ
from the student-generated prefixes encountered at inference
\citep{agarwal2024policy}. \emph{On-policy} distillation reduces this mismatch
by applying the distillation loss to trajectories sampled from the student
itself, so that the model is supervised on its own inference-time states
\citep{agarwal2024policy}.

In the on-policy self-distillation (OPSD) instantiation, the teacher and student share
the same model, so no external teacher is required \citep{zhao2026self}.
Given a multimodal input $(q,\mathcal{I})$, we first sample a response
$y \sim \pi_\theta(\cdot \mid q,\mathcal{I})$ and then align the student's
per-token distribution with a teacher distribution defined over the same model:
\begin{equation}
\begin{split}
\mathcal{L}_{\mathrm{OPSD}}(\theta)
= \mathbb{E}_{(q,\mathcal{I})}\,
  \mathbb{E}_{y\sim\pi_\theta(\cdot\mid q,\mathcal{I})}
  \Bigg[
  \frac{1}{L}\sum_{k=1}^{L}
  D_{\mathrm{KL}}\Big(
      \pi_{\mathrm{teacher}}(\cdot \mid y_{<k}, q, \mathcal{I})
      \,\big\|\,
      \pi_\theta(\cdot \mid y_{<k}, q, \mathcal{I})
  \Big)
  \Bigg].
\end{split}
\end{equation}
Here the gradient is taken with respect to the student parameters $\theta$,
while the teacher distribution is treated as a fixed target. Sampling
$y$ from $\pi_\theta$ makes the objective on-policy, and defining
$\pi_{\mathrm{teacher}}$ from the same model makes it self-distillation.

Our method adopts this OPSD view only as a diagnostic principle: rather than
optimizing a distillation objective, we replay the same model under two
related contexts and use the resulting distributional shift to estimate whether
a historical reasoning segment still contains information not covered by a
compact task-state summary.

\noindent\textbf{Attention over modalities.}
For visualization, we measure text-to-image attention across decoder layers following prior work \citep{chen2024image}. The exact definition is provided in Appendix~\ref{app:preliminaries_extra}.

\section{Method}
\label{sec:method}


We propose \textbf{\textsc{SPARE}} (Selective Pruning of Accumulated Reasoning with Visual Evidence Preservation), a
post-hoc context-pruning method for multi-step MLLM agents, as shown in Figure \ref{fig:spare_overview}. The motivation is
that explicit reasoning is useful for decomposing visual questions, planning
tool calls, and integrating intermediate observations, but retaining the entire
self-generated textual trace indefinitely can create \emph{textual debt}: the
agent's later predictions become increasingly conditioned on its own textual
history rather than on the visual evidence underlying the task. \textsc{SPARE}
therefore does not aim to suppress reasoning generation. Instead, it diagnoses
which accumulated text has low residual sensitivity to a compact task state and
compresses only those low-residue segments, while preserving modality-critical
content such as OCR strings, coordinates, bounding boxes, visual anchors, tool
calls, and external observations. In this sense, \textsc{SPARE} compresses
linguistic scaffolding rather than task state.

\subsection{Summary-Conditioned Coverage Estimation}
\label{sec:summary-conditioned-coverage}

\paragraph{Adaptive summary trigger.}

\textsc{SPARE} is invoked only after the student agent itself calls an internal
\texttt{summarize\_the\_task} tool and produces a compact task-state summary
$\sigma$. This self-generated summary contain the original question, completed
tool calls, accumulated visual evidence, and remaining uncertainty. We use
$\sigma$ only as auxiliary diagnostic context to test which prior reasoning
spans are already covered by the consolidated task state. This adaptive trigger
avoids fixed-interval compression: short trajectories incur no pruning cost, and
longer trajectories are considered for pruning only after the student has
explicitly produced a compact state.

\paragraph{Summary-conditioned replay.}
Let $\boldsymbol{x}=(x_1,\ldots,x_N)$ be the tokenized concatenation of candidate
reasoning segments, separated by a fixed delimiter. To test whether the
self-generated summary $\sigma$ covers a segment's information, we replay
$\boldsymbol{x}$ under two contexts:
\begin{align}
    \rho_{\mathrm{stu}} &= \rho_0, \\
    \rho_{\mathrm{tea}} &= \rho_0 \,\Vert\, \sigma,
\end{align}
where $\rho_0$ contains the shared system, tool-use, and task prefix. The
teacher context receives $\sigma$ as privileged information, while the student
context does not. Importantly, this does not introduce an external teacher
model: the teacher distribution is produced by the same model, conditioned on
the self-generated summary. The same model $\pi_\theta$ is then queried under both
contexts:
\begin{align}
    p_k^{\mathrm{stu}}(u)
    &= P_{\pi_\theta}\!\left(u \mid \rho_{\mathrm{stu}}, \boldsymbol{x}_{<k}\right), \\
    p_k^{\mathrm{tea}}(u)
    &= P_{\pi_\theta}\!\left(u \mid \rho_{\mathrm{tea}}, \boldsymbol{x}_{<k}\right),
\end{align}
where $k$ indexes replay tokens and $u$ indexes vocabulary tokens. If adding
$\sigma$ changes the continuation distribution only slightly, the summary
already covers the replayed content. A larger shift indicates that the original reasoning contains information not represented in the summary and should therefore be preserved.

\paragraph{Top-$K$ reverse-KL coverage score.}
For each replay token position $k$, we compute a truncated reverse-KL score over
the student's top-$K$ vocabulary support $\Omega_k^{\mathrm{stu}}$:
\begin{equation}
\label{eq:spare-rkl}
    d_k
    =
    \left[
    \sum_{u\in\Omega^{\mathrm{stu}}_k}
    p_k^{\mathrm{stu}}(u)
    \left(
        \log p_k^{\mathrm{stu}}(u)
        -
        \log p_k^{\mathrm{tea}}(u)
    \right)
    \right] .
\end{equation}
We use $K=20$ in all experiments. If $u\in\Omega_k^{\mathrm{stu}}$ is absent
from the teacher top-$K$ support, $\log p_k^{\mathrm{tea}}(u)$ is set to $-20$.
The resulting score $d_k$ measures the residual information not covered by the
summary: high values indicate that the token remains summary-sensitive and
should be protected, while low values indicate that the summary sufficiently
explains the token's context.

 KL scales vary across pruning events, we normalize scores within each
event:
\begin{equation}
\label{eq:spare-norm}
    \tilde d_k
    =
    \frac{d_k - d_{\min}}
         {d_{\max} - d_{\min} + \epsilon},
    d_{\min}=\min_{k'} d_{k'},
    d_{\max}=\max_{k'} d_{k'} .
\end{equation}
We set $\epsilon=10^{-12}$ and assign all normalized scores to zero when
$d_{\max}=d_{\min}$. Segment-level pruning then aggregates $\tilde d_k$ to
determine whether the summary covers each reasoning segment.

\subsection{Segment-Level Pruning}
\label{sec:segment-pruning}

\paragraph{From token scores to segment decisions.}
The replay sequence $\boldsymbol{x}$ is formed by concatenating historical
assistant reasoning segments. Let $h_i$ be the $i$-th candidate segment and
\begin{equation}
    \mathcal{P}_i = \{k : x_k \text{ belongs to } h_i\}
\end{equation}
denote its token positions in the replay sequence. Since segment boundaries
are recorded during concatenation, the normalized token-level scores
$\tilde d_k$ can be mapped back to their original reasoning segments. A segment is selected for evidence preservation if it contains at least
$\kappa$ high-sensitivity tokens:
\begin{equation}
\label{eq:spare-keep}
    \mathrm{Preserve}(h_i)
    =
    \mathbf{1}\left\{
        \left|\left\{k\in\mathcal{P}_i:
        \tilde d_k > \tau\right\}\right|
        \ge \kappa
    \right\}.
\end{equation}
High-KL tokens indicate information that is not sufficiently covered by the
task-state summary. Segments containing enough such tokens are therefore
routed to evidence-preserving reconstruction. For low-residue segments, the
reasoning text is removed entirely because its information is already
represented in the summary.

\paragraph{Why a count-based threshold?}
We use a count rule rather than an average score because task-critical evidence
is often sparse. A long reasoning segment may be mostly redundant while still
containing a few crucial tokens, such as an OCR string, coordinate, bounding
box, or candidate label. Averaging can dilute these localized signals, whereas
the count rule identifies segments containing sparse but potentially important
evidence.

\subsection{Visual Evidence Preservation}
\label{sec:evidence-preservation}

\paragraph{KL-driven reconstruction.}
The segment-level decision determines how each reasoning span is compressed.
For high-KL segments, \textsc{SPARE} replaces the original lengthy reasoning
with concise, verifiable visual evidence, such as crop locations, recognized
text, numeric values, or relative relations among candidates. This preserves
information not adequately represented in the summary without retaining the
full reasoning span. For low-KL segments, the reasoning text is removed
entirely. In both cases, original tool actions, tool outputs, visual
observations, and image tokens remain unchanged.

\paragraph{Role of the task-state summary.}
The task-state summary is used as auxiliary context for the next reasoning
step, but is not itself written into the assistant reasoning trace or used as
a direct replacement for historical segments. It serves two purposes: it
provides the privileged context used to estimate which information remains
uncovered, and it carries the consolidated task state forward after pruning.
High-KL visual evidence is retained alongside this summary because the KL
score indicates that the summary alone is insufficient.

\subsection{Strengthening the Summarizer via SFT}
\label{sec:sft-summarizer}
The effectiveness of \textsc{SPARE} depends on the quality of the task-state summary: a summary that covers more of the accumulated reasoning enables more aggressive low-residue pruning without disrupting future decisions. To this end, we further fine-tune the summarizer via supervised fine-tuning (SFT). A stronger summarizer expands the coverage of each summary, so that a larger fraction of reasoning segments can be safely explained by the summary alone. Under the same reverse-KL criterion, more segments therefore fall into the low-residue regime and become prunable, while high-residue visually-grounded segments remain preserved.

\section{Experiments}
\label{sec:experiments}

\subsection{Experimental Settings}
\label{ssec:Experimental_Settings}

\paragraph{Benchmarks.}
We evaluate \textsc{SPARE} on several visual tool-use benchmarks requiring multi-step reasoning, tool invocation, and intermediate visual observations. \textbf{VisualToolBench} (VTB)~\citep{guo2025beyond} requires models to actively ``think with images'' through operations such as cropping, editing, and enhancement,
a paradigm increasingly studied in multimodal reasoning
\citep{su2025thinkingimagesmultimodalreasoning}. \textbf{m\&m's} (MNMS)~\citep{ma2024m} includes 4000+ multi-step multimodal tasks over 33 tools, covering multimodal models, public APIs, and image-processing modules. \textbf{GTA}~\citep{wang2024gta} provides real-world tasks with human-written queries and executable tool chains across perception, operation, logic, and creativity. \textbf{V$^{\ast}$}~\citep{wu2024v} evaluates guided visual search on high-resolution images, requiring fine-grained target localization before answering. \textbf{BLINK-Jigsaw} (B-Jig.)~\citep{fu2024blink} tests spatial perception by requiring models to reassemble image fragments. Together, these benchmarks cover long-horizon tool use and perception-intensive reasoning, making them suitable for testing whether reasoning compression preserves visual grounding.

\paragraph{Backbones.}
To evaluate SPARE's generality, we test three vision-language backbones:
Qwen3-VL-8B-Instruct,
Qwen3-VL-30B-A3B-Instruct, and Llama-3.1-Nemotron-Nano-VL-8B-V1.
We keep generation settings and the tool interface fixed.

\paragraph{Compared methods.}
For each backbone, our main comparison includes five inference strategies
under an identical tool-use harness.
(i) \textbf{Tool Baseline (Full Trace)} executes the standard multi-round
tool loop and retains the complete reasoning-action-observation history.
(ii) \textbf{- Tools (Direct Answer)} answers the query directly without
invoking tools and serves as a non-agentic reference.
(iii) \textbf{+ Delete All Reasoning} removes every completed reasoning span
while preserving the original action blocks, tool observations, and images.
(iv) \textbf{+ Visual Evidence-Only} non-selectively rewrites every completed
reasoning span as compact structured visual evidence while preserving its
original action block.
(v) \textbf{+ SPARE (Ours)} uses summary-conditioned KL to estimate the
residual information in each reasoning span, removing low-residue reasoning
and reconstructing high-residue reasoning as structured visual evidence. 
Full Trace provides the complete-context reference, while Delete All Reasoning
and Visual Evidence-Only isolate the effects of non-selective deletion and
evidence reconstruction, respectively. Additional random and KL-guided controls used for the accuracy-pruning
analysis are defined in Appendix~\ref{app:accuracy_pruning_tradeoff}.

\paragraph{Metrics.}
Following each benchmark’s official protocol, we report task success or accuracy, where higher is better ($\uparrow$). We also report \textbf{Pruned (\%)}, the net percentage of reasoning-side history tokens removed relative to Tool Baseline. Token counts include remaining reasoning, structured evidence blocks, and any task-state summary, but exclude unchanged action blocks, tool observations, and image tokens. Thus, Full Trace has $0.0\%$ pruning, Delete All Reasoning has $100.0\%$, and Direct Answer is assigned $0.0\%$ for reporting consistency as it produces no comparable trajectory. The main table reports the macro-average pruning ratio across benchmarks on the common evaluation set.

\paragraph{Implementation details.}
\textsc{SPARE} is applied purely at test time. Pruning is triggered only when the model invokes the internal \texttt{summarize\_the\_task} tool, so short trajectories without summaries incur no compression cost. We compute the truncated reverse-KL coverage score on the student’s top-$K$ support and map token scores to reasoning segments using the count-based rule with fixed $\tau=0.2$ and $\kappa=2$ across all models and benchmarks (Section~\ref{sec:method}). The same model provides both student and teacher distributions by conditioning on prompts with and without the task-state summary $\sigma$, avoiding any auxiliary model.

\subsection{Main Results}
\label{sec:main_results}

\input{tables/main_results}

Table~\ref{tab:main_results} reports task accuracy and reasoning pruning
across five benchmarks. \textsc{SPARE} achieves the highest average accuracy
among the pruning methods for all three backbones, while the non-selective
Delete All Reasoning and Visual Evidence-Only baselines consistently reduce
overall performance. On Qwen3-VL-30B, \textsc{SPARE} improves average
accuracy from 44.30\% to 49.19\% while pruning 63.70\% of reasoning, with
particularly strong gains on GTA, V$^\ast$, and MNMS. On Qwen3-VL-8B, it
maintains comparable performance (53.47\% vs.\ 54.27\%) with 37.89\% pruning
and improves Full Trace on GTA and V$^\ast$.

Nemotron shows a different pattern: Direct Answer outperforms Full Trace on
all applicable benchmarks, suggesting that accumulated tool-use context can
interfere with later decisions when tool-use capability is weaker. Even so,
\textsc{SPARE} improves over Full Trace on GTA, B-Jigsaw, and VTB and nearly
matches its average accuracy (29.97\% vs.\ 30.10\%) while pruning 64.58\% of
reasoning. Overall, these results show that selective pruning with
visual-evidence preservation provides a better accuracy--compression balance
than uniformly deleting or reconstructing the reasoning history.


\paragraph{Accuracy--pruning trade-off.}

\begin{figure}[t]
  \centering
  \includegraphics[width=0.8\linewidth]
  {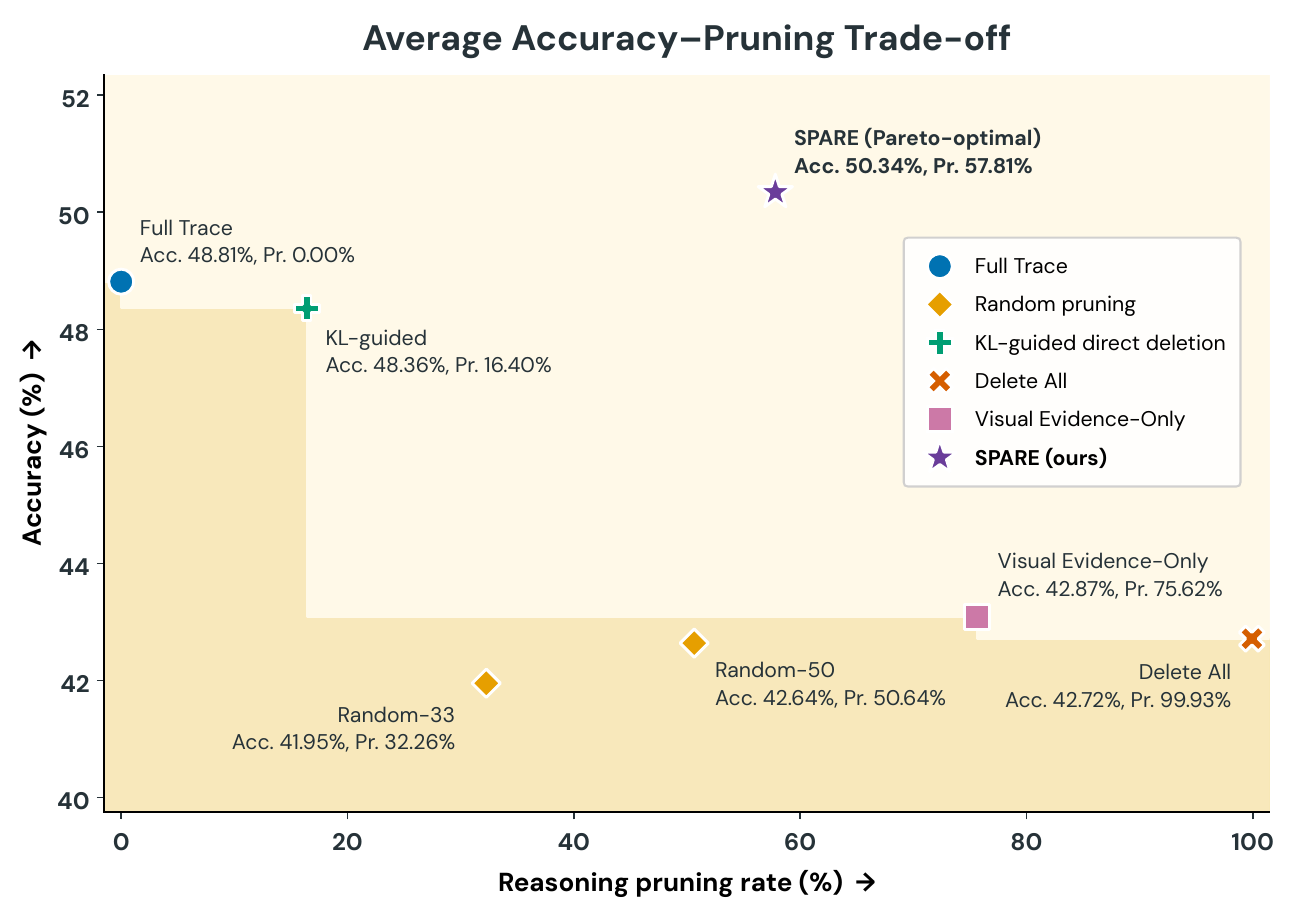}
  \caption{Accuracy--pruning trade-off averaged across three models on
  BLINK-Jigsaw, GTA, and MNMS. Accuracy and reasoning-pruning rates are
  weighted by the number of evaluation samples. The staircase connects the
  Pareto-optimal comparison methods, and the upper-right region indicates a
  better trade-off. SPARE lies above this frontier, attaining 50.34\% accuracy
  while pruning 57.81\% of reasoning tokens.}
  \label{fig:accuracy_pruning_tradeoff}
\end{figure}

Figure~\ref{fig:accuracy_pruning_tradeoff} compares the aggregate
accuracy--pruning trade-off across three models on BLINK-Jigsaw, GTA, and
MNMS. Indiscriminate pruning substantially reduces accuracy: aggregate
accuracy drops from 48.81\% with Full Trace to 41.95\% with Random-33 and
remains near 42\% under more aggressive random or complete deletion.
KL-guided pruning recovers 48.36\% accuracy but removes only 16.40\% of
reasoning, while Visual Evidence-Only achieves 42.86\% accuracy with 75.62\%
pruning. In contrast, \textsc{SPARE} achieves the highest accuracy of 50.34\%
while pruning 57.81\%, placing it above the baseline Pareto frontier. These
results show that KL-based selection identifies which reasoning remains
important, while evidence reconstruction preserves that information
compactly. Complete per-model results are reported in
Table~\ref{tab:complete_accuracy_pruning} in
Appendix~\ref{app:accuracy_pruning_tradeoff}.

\paragraph{Textual interference and visual attention.}
On a controlled subset where early reasoning conflicts with later visual or
tool evidence, \textsc{SPARE} improves accuracy by 27.27 percentage points
and reduces copying of outdated conclusions by the same margin for both
Qwen3-VL backbones. Random deletion produces smaller gains, showing that the
improvement cannot be explained by context shortening alone. In a separate
analysis of 10 tool-use trajectories, pruning also increases attention to
image tokens across nearly all decoder layers. Together, these observations
support the hypothesis that selective pruning reduces interference from
accumulated reasoning and strengthens reliance on visual evidence
(Appendix~\ref{app:supp_analysis}).

\subsection{SFT for Summarization Capability}
\label{sec:sft_results}

\begin{wraptable}{r}{0.52\columnwidth}
\vspace{-6pt}
\centering
\small
\caption{SFT results on VTB and MNMS.}
\label{tab:sft-summary-results}
\vspace{4pt}
\setlength{\tabcolsep}{4pt}

\begin{tabularx}{\linewidth}{@{}Xcc@{}}
\toprule
\textbf{Model / Metric} & \textbf{VTB} & \textbf{MNMS} \\
\midrule
Base (Qwen3-VL-8B)
& 26.83 & 60.61 \\
+ SFT summarizer
& \textbf{28.73} & \textbf{65.44} \\
\midrule
More pruned tokens vs.\ Base
& \textbf{23.8\%} & \textbf{13.5\%} \\
\bottomrule
\end{tabularx}
\vspace{-6pt}
\end{wraptable}

We further study whether stronger task-state summaries enable more aggressive
pruning. To this end, we collect 376 on-policy trajectories from
Qwen3-VL-8B on VTC-Bench (VTC)~\citep{zhu2026vtc} and prompt the stronger
Qwen3-235B to produce a compact task-state summary at each intermediate
context. Qwen3-VL-8B is then fine-tuned on the resulting (context, summary)
pairs with a standard next-token objective, distilling the 235B model's
summarization capability into the 8B agent. Table~\ref{tab:sft-summary-results} reports task accuracy under the
\textsc{SPARE} pipeline before and after SFT-based summarizer fine-tuning. The
last row shows the relative increase in pruned tokens per trajectory compared
with the base summarizer. The SFT summarizer enables more aggressive pruning
and improves task accuracy on both VisualToolBench and MNMS, supporting our
motivation that stronger summaries expand the safe-to-prune region and allow
more redundant reasoning to be removed without harming downstream decisions.

\subsection{Ablation Study}
\label{ssec:ablation}

\begin{wraptable}{r}{0.52\columnwidth}
\vspace{-6pt}
\centering
\small
\caption{Ablations on GTA. Accuracy and pruning rates are percentages.}
\label{tab:ablation}
\vspace{4pt}
\setlength{\tabcolsep}{4pt}

\begin{tabularx}{\linewidth}{@{}Xcc@{}}
\toprule
\textbf{Ablation Setting}
& \textbf{Acc.}
& \textbf{Pr.} \\

\midrule
\multicolumn{3}{@{}l@{}}{
\begin{tabular}[c]{@{}l@{}}
\textit{Selective Pruning and}\\
\textit{Evidence Reconstruction}
\end{tabular}
} \\
Full trace baseline
& 40.68 & 0.00 \\
Summary probe only
& 35.59 & 0.00 \\
${}+{}$ KL selection
& 40.68 & 0.00 \\
${}+{}$ Evidence
& 37.29 & 56.08 \\
\rowcolor{spareBg}
\textbf{${}+{}$ KL selection ${}+{}$ Evidence (ours)}
& \textbf{42.37} & \textbf{62.42} \\

\midrule
\multicolumn{3}{@{}l@{}}{\textit{Adaptive Summary Triggering}} \\
Every round
& 32.20 & 52.54 \\
Before final answer
& 38.98 & 51.38 \\
\rowcolor{spareBg}
\textbf{Model-selected (ours)}
& \textbf{42.37} & \textbf{62.42} \\

\midrule
\multicolumn{3}{@{}l@{}}{
\textit{Parameter Robustness} ($\tau=0.2$)
} \\
$\kappa=1$
& \textbf{42.37} & 62.42 \\
\rowcolor{spareBg}
$\boldsymbol{\kappa=2}$ \textbf{(default)}
& \textbf{42.37} & 62.42 \\
$\kappa=3$
& 38.98 & \textbf{63.16} \\

\bottomrule
\end{tabularx}
\vspace{-6pt}
\end{wraptable}

\paragraph{Why combine selective pruning with evidence reconstruction?}
We conduct ablations using Llama-3.1-Nemotron-Nano-VL-8B-V1 on GTA, with the results summarized
in Table~\ref{tab:ablation}. 
\textbf{Full trace} retains all reasoning as the reference.
\textbf{Summary probe only} summarizes the current task state to assist
subsequent reasoning but does not prune the history.
\textbf{+ KL selection} deletes low-KL segments while retaining high-KL
segments in their original form.
\textbf{+ Evidence} omits KL selection and replaces every reasoning segment
with structured visual evidence.
The complete \textsc{SPARE} removes low-KL segments and reconstructs high-KL
segments as concise visual evidence. The results show that summary assistance alone is insufficient. KL selection avoids unsafe pruning but provides no compression when high-KL reasoning is retained verbatim, whereas evidence-only reconstruction compresses the history at the cost of accuracy. Combining selection and reconstruction achieves the
best accuracy (42.37\%) with substantial pruning (62.42\%), demonstrating that the two components are complementary.

\paragraph{When should the context be summarized and pruned?}
The second block compares three triggering policies.
\textbf{Every round} forces the model to summarize after each reasoning round,
whereas \textbf{Before final answer} invokes the summary once immediately
before answering. \textbf{Model-selected} allows the model to decide whether
and when summarization is necessary. Frequent forced summarization can disrupt
reasoning, while summarizing only before the final answer cannot control
earlier context accumulation. The model-selected policy achieves the highest
accuracy (42.37\%) and pruning rate (62.42\%), demonstrating that adaptive
summarization is more effective than fixed triggering schedules. We next ablate the KL-based decision rule by varying the required number of consecutive pruning decisions, $\kappa\in{1,2,3}$, while fixing $\tau=0.2$. The settings $\kappa=1$ and $\kappa=2$ yield identical results. Increasing $\kappa$ to 3 slightly improves the pruning rate but reduces accuracy. Based on this local stability, we adopt $(\tau,\kappa)=(0.2,2)$ as the default setting.

\section{Related Work}
\label{sec:Related}
\paragraph{Token Pruning for MLLMs}
Existing MLLM token pruning methods primarily remove redundancy from the visual stream \citep{chen2024image,yang2025visionzip,shang2025llava,tan2025tokencarve,zhang2025beyond,lee2025tamp}. While effective for efficiency, this objective can be misaligned with multimodal reasoning, where questions are often partially answerable from textual cues and visual evidence is already under-attended in deep layers \citep{zhang2025adaptinfer,zhang2025beyond,zhang2026chain,takezoe2026learnpruner}. Concise-reasoning and generic summarization methods can shorten generated text or compress history, but do not diagnose which historical segments remain functionally necessary. Our method instead targets accumulated \emph{textual} redundancy in tool-use trajectories and prunes only segments already covered by a compact task-state summary.

\paragraph{Improving Visual Grounding}
Weak visual grounding can arise both from insufficient visual perception and from language-prior bias, where accumulated textual context overwhelms otherwise available visual evidence. Prior work addresses the latter through counterfactual training \citep{niu2021counterfactual}, contrastive decoding \citep{zhao2025cross}, attention calibration \citep{woo2025don,fazli2025mitigating,zhao2025looking}, and preference optimization \citep{xie2024v,chaubey2026mod}, but leaves the competing textual context intact. A complementary line strengthens perception itself. Thinking-with-images methods actively manipulate or construct intermediate images during reasoning \citep{imageofthought,su2025thinkingimagesmultimodalreasoning}, while Vision-OPD and Visual-OPSD distill signals from privileged visual evidence or visual thoughts \citep{visionopd,visualopsd}. SPARE instead targets the context-side cause of weak grounding by repurposing OPSD-style policy divergence as a diagnostic for pruning stale self-generated text, thereby preserving the original visual evidence rather than training new visual capabilities.

\section{Conclusion}
\label{sec:conclusion}

In this paper, we identify \emph{textual debt} as a key failure mode of multi-step MLLM agents, where accumulated self-generated reasoning gradually dominates the context and weakens reliance on visual evidence. Our motivation is that pruning reasoning tokens can help because their utility changes over time: once task-relevant visual information has been captured, later text often becomes redundant, while incorrect or incomplete early grounding can make accumulated text reinforce stale linguistic assumptions. Based on this insight, we propose \textsc{SPARE}, a OPSD KL-guided framework that selectively removes redundant reasoning while preserving visual evidence, thereby reducing textual dominance and redirecting attention to images. Experiments show that \textsc{SPARE} improves task performance, reduces reasoning-token usage, and restores attention to visual evidence, suggesting that inference-time selective forgetting is an effective mechanism for long-context multimodal reasoning.

\bibliographystyle{plainnat}
\bibliography{reference}


\appendix
\input{sections/appendix}



\end{document}

%% file: tables/main_results.tex
\definecolor{spareBg}{RGB}{254, 243, 232}
\newcolumntype{Y}{>{\centering\arraybackslash}X}

\begin{table*}[t]
\centering
\caption{Task performance and average reasoning pruning across five visual
tool-use benchmarks. Benchmark columns report accuracy. Acc. and Pr. denote average accuracy and reasoning-token pruning rate,
respectively. Note that MNMS evaluates
tool planning and therefore is inapplicable for Direct Answer mode.}
{\small

\begin{tabularx}{\textwidth}{
>{\raggedright\arraybackslash}p{0.26\textwidth}
Y Y >{\centering\arraybackslash}p{0.08\textwidth} Y >{\centering\arraybackslash}p{0.09\textwidth} Y Y
}

\toprule

\textbf{Model / Metric}
& \textbf{GTA} $\uparrow$
& \textbf{V$^\ast$} $\uparrow$
& \textbf{B-Jig.} $\uparrow$
& \textbf{VTB} $\uparrow$
& \textbf{MNMS} $\uparrow$
& \textbf{Acc.} $\uparrow$
& \textbf{Pr.} $\%$
\\

\midrule


\multicolumn{8}{@{}l}{\textit{Qwen3-VL-30B-A3B-Instruct}}
\\[1pt]

Tool Baseline (Full Trace)
& 45.76 & 55.85 & \underline{67.33}
& \textbf{27.31} & \underline{36.84} & \underline{44.30} & 0.00
\\

\quad $-$ Tools (Direct Answer)
& 33.90 & 52.36 & \textbf{70.67}
& 21.43 & -- & 43.42 & 0.00
\\

\quad + Delete All Reasoning
& 50.85 & 37.17 & 57.33
& 26.05 & 24.56 & 35.22 & 99.86
\\

\quad + Visual Evidence-Only
& \textbf{59.32} & \underline{61.78} & 64.00
& \underline{26.47} & 25.44 & 42.73 & 77.09
\\

\rowcolor{spareBg}
\quad + \textbf{SPARE (ours)}
& \textbf{59.32} & \textbf{62.30} & 64.00
& \underline{26.47} & \textbf{49.56} & \textbf{49.19} & 63.70
\\

\midrule


\multicolumn{8}{@{}l}{\textit{Qwen3-VL-8B-Instruct}}
\\[1pt]

Tool Baseline (Full Trace)
& 42.37 & \underline{72.77} & \textbf{74.00}
& \underline{23.53} & \textbf{60.96} & \textbf{54.27} & 0.00
\\

\quad $-$ Tools (Direct Answer)
& 42.37 & 63.35 & \underline{72.00}
& 20.17 & -- & 47.34 & 0.00
\\

\quad + Delete All Reasoning
& 40.68 & 68.59 & 62.00
& \underline{23.53} & 57.02 & 50.12 & 100.00
\\

\quad + Visual Evidence-Only
& \underline{44.07} & 69.63 & 59.33
& \textbf{24.79} & 56.58 & 50.35 & 72.58
\\

\rowcolor{spareBg}
\quad + \textbf{SPARE (ours)}
& \textbf{49.15} & \textbf{73.30} & \underline{72.00}
& 23.11 & \underline{57.46} & \underline{53.47} & 37.89
\\

\midrule


\multicolumn{8}{@{}l}{%
\textit{Llama-3.1-Nemotron-Nano-VL-8B-V1}}
\\[1pt]

Tool Baseline (Full Trace)
& 40.68 & \underline{48.69} & 45.33
& 6.16 & \textbf{26.75} & \underline{30.10} & 0.00
\\

\quad $-$ Tools (Direct Answer)
& \textbf{44.07} & \textbf{56.02} & \textbf{51.33}
& \textbf{9.46} & -- & \textbf{36.44} & 0.00
\\

\quad + Delete All Reasoning
& 37.29 & 45.03 & 46.67
& 6.75 & 21.49 & 28.07 & 99.38
\\

\quad + Visual Evidence-Only
& 37.29 & 43.98 & 44.00
& 6.26 & 17.98 & 26.32 & 54.84
\\

\rowcolor{spareBg}
\quad + \textbf{SPARE (ours)}
& \underline{42.37} & 46.60 & \underline{48.67}
& \textbf{9.46} & \underline{21.93} & 29.97 & 64.58
\\

\bottomrule
\end{tabularx}
}
\label{tab:main_results}
\end{table*}

%% file: sections/appendix.tex
\section{Additional Preliminaries}
\label{app:preliminaries_extra}
\subsection{Multi-Step Tool Use in MLLMs}
\label{ssec:prelim_setup}
We consider a vision language model (VLM) $\pi_\theta$ that answers a user
question through multi-step, tool-augmented reasoning. The input consists of a
question $q$ and a set of image tokens $\mathcal{I}$, and the model may query a
fixed set of external tools $\mathcal{T}$ over at most $T$ interaction steps.

At step $t$, the model conditions on the current message history
$\mathbf{H}_t$ and produces an assistant message
\begin{equation}
    m_t = (h_t, a_t),
\end{equation}
which is composed of a free-form reasoning span $h_t$ and an action block
$a_t$. The action block is either a tool invocation
$\langle\text{tool\_call}\rangle$ or a final answer
$\langle\text{response}\rangle$. If $a_t$ invokes a tool, the corresponding
tool result is appended to the history, yielding $\mathbf{H}_{t+1}$, and the
interaction proceeds to the next step; if $a_t$ emits a final response, the
episode terminates. We denote the sequence of past reasoning--action segments
by $\mathcal{S}_t = \{\,s_i = (h_i, a_i)\,\}_{i<t}$.

\subsection{On-policy self-distillation.}
Knowledge distillation trains a student model $\pi_\theta$ to match a teacher distribution $\pi_{\mathrm{teacher}}$ over next tokens. Conventional off-policy distillation minimizes this divergence on fixed teacher or ground-truth sequences, which creates exposure bias because training prefixes differ from the student-generated prefixes encountered at inference \citep{agarwal2024policy}. \emph{On-policy} distillation reduces this mismatch by applying the distillation loss to trajectories sampled from the student itself, so the model is supervised on its own inference-time states \citep{agarwal2024policy}.

In the on-policy self-distillation instantiation, the teacher and the student share
the same model, so no external teacher is required \citep{zhao2026self}. Concretely, given a
multimodal input $(q, \mathcal{I})$, we first roll out a response on-policy,
$y \sim \pi_\theta(\cdot \mid q, \mathcal{I})$, and then align the student's per-token
distribution with a teacher distribution $\pi_{\mathrm{teacher}}$ defined over
the same model:
\begin{equation}
\begin{split}
  \mathcal{L}_{\mathrm{OPSD}}(\theta)
  = \mathbb{E}_{(q,\mathcal{I})}\,
    \mathbb{E}_{y \sim \pi_\theta(\cdot \mid q, \mathcal{I})}
    \Bigg[\frac{1}{L}\sum_{k=1}^{L}  \\
      D_{\mathrm{KL}}\Big(
        \pi_{\mathrm{teacher}}(\cdot \mid y_{<k}, q, \mathcal{I})
       \,\big\|\,
        \pi_\theta(\cdot \mid y_{<k}, q, \mathcal{I})
      \Big)\Bigg],
\end{split}
\end{equation}
where the gradient is taken with respect to the student parameters $\theta$ and
the teacher distribution is treated as a fixed target (stop-gradient). Sampling
$y$ from $\pi_\theta$ makes the objective on-policy, while defining
$\pi_{\mathrm{teacher}}$ from the same model makes it self-distillation. This
formulation lets the model learn from its own generated trajectories and provides the training signal on which our method builds.

\subsection{Self-attention over modalities.}
Within a transformer layer $\ell$ with attention heads indexed by $b$, the
attention weight from a query token $i$ to a key token $j$ is
\begin{equation}
  A^{(\ell,b)}_{ij}
  = \mathrm{softmax}_j\!\left(
      \frac{q^{(\ell,b)}_i \cdot k^{(\ell,b)}_j}{\sqrt{d}}
    \right),
  \qquad \sum_{j=1}^{N} A^{(\ell,b)}_{ij} = 1,
\end{equation}
where $q$ and $k$ are the query and key projections and $d$ is the head
dimension. For a query token $i$ we measure how much of its attention is directed
to the visual stream by summing over visual keys and averaging over heads,
\begin{equation}
  a^{(\ell)}_{i \to \mathcal{I}}
  = \frac{1}{H}\sum_{b=1}^{H}\sum_{j \in \mathcal{I}} A^{(\ell,b)}_{ij},
  \label{eq:visual_attention}
\end{equation}
which we refer to as the visual attention ratio of token $i$ at layer
$\ell$. Aggregating $a^{(\ell)}_{i \to \mathcal{I}}$ over the textual query tokens
gives a scalar summary of how strongly the model attends to the image while
producing its response. A well-documented empirical observation is that this
ratio is high in the first few layers but decays sharply with depth, so that in
deep layers the model attends almost entirely to textual tokens
\citep{chen2024image}.

\section{Additional Experimental Details}
\label{app:exp_details}

\subsection{Reasoning-Token Accounting}
All tool-based methods use the same backbone, decoding configuration, tool
set, and execution environment. They differ only in how accumulated
assistant reasoning is retained or reconstructed. Original tool calls,
tool observations, and images remain unchanged. Task-state summaries used
by SPARE are transient probes and are not written into the persistent
trajectory.

\paragraph{Counting scope.}
For method $m$, let $\mathcal{D}_b$ denote the evaluated examples from
benchmark $b$, and let $\mathcal{H}^{m}_{j,b}$ denote all completed tool-use
reasoning segments generated for example $j\in\mathcal{D}_b$.

For each segment $h\in\mathcal{H}^{m}_{j,b}$,
$R_{\mathrm{orig}}(h)$ is the original assistant reasoning preceding its
corresponding \texttt{<tool\_call>} block, and
$\widetilde{R}^{m}_{\mathrm{final}}(h)$ is its final persistent
reasoning-side representation. The latter equals the original reasoning for
an unpruned segment, is empty when the reasoning is deleted, and contains
the reconstructed structured visual evidence when the segment is compressed.

Tool-call JSON, tool observations, images, final answers, and transient
summary probes are excluded. All token counts use the tokenizer and chat
serialization of the evaluated backbone.

\paragraph{Multiple summary events.}
A trajectory may invoke the summary tool multiple times. Each reasoning
segment is counted exactly once using its original text and final persistent
representation, regardless of how many summary events inspect or modify it.

\paragraph{Original and retained reasoning tokens.}
The total number of original reasoning tokens generated by method $m$ on
benchmark $b$ is
\begin{equation}
O_{m,b}
=
\sum_{j\in\mathcal{D}_b}
\sum_{h\in\mathcal{H}^{m}_{j,b}}
\operatorname{Tok}\!\left(R_{\mathrm{orig}}(h)\right).
\label{eq:original_reasoning_tokens}
\end{equation}

The number of reasoning tokens remaining in the persistent history is
\begin{equation}
K_{m,b}
=
\sum_{j\in\mathcal{D}_b}
\sum_{h\in\mathcal{H}^{m}_{j,b}}
\operatorname{Tok}\!\left(
\widetilde{R}^{m}_{\mathrm{final}}(h)
\right).
\label{eq:retained_reasoning_tokens}
\end{equation}

\paragraph{Removed reasoning tokens.}
The total number of net reasoning tokens removed is
\begin{equation}
P_{m,b}
=
O_{m,b}-K_{m,b}.
\label{eq:removed_reasoning_tokens}
\end{equation}
The difference is not clipped if an evidence reconstruction is longer than
its original reasoning segment.

\paragraph{Reasoning-token pruning rate.}
The reasoning-token pruning rate reported in the main table is
\begin{equation}
r_{m,b}
=
100
\times
\frac{O_{m,b}-K_{m,b}}{O_{m,b}}.
\label{eq:reasoning_pruning_rate}
\end{equation}
Higher values indicate stronger compression of the persistent reasoning
history. The rate is computed from aggregate token counts rather than by
averaging per-example percentages.

Full Trace has $r_{m,b}=0\%$. Direct Answer contains no comparable tool-use
reasoning history and is therefore reported as N/A.

\subsection{Accuracy-Pruning Trade-off}
\label{app:accuracy_pruning_tradeoff}


\begin{table*}[t]
\centering
\caption{
Complete per-model results on the three benchmarks.
Acc. denotes accuracy, and Pr. denotes the reasoning-pruning rate.
All values are percentages.
Figure~\ref{fig:accuracy_pruning_tradeoff} aggregates the results using
benchmark-size weighting: 150 samples for BLINK-Jigsaw,
59 answer-scored samples for GTA, and 228 samples for MNMS.
}
\label{tab:complete_accuracy_pruning}

\setlength{\tabcolsep}{4.0pt}
\renewcommand{\arraystretch}{1.08}

\resizebox{\textwidth}{!}{%
\begin{tabular}{@{}llrrrrrr@{}}
\toprule
& &
\multicolumn{2}{c}{BLINK-Jigsaw} &
\multicolumn{2}{c}{GTA} &
\multicolumn{2}{c}{MNMS} \\
\cmidrule(lr){3-4}
\cmidrule(lr){5-6}
\cmidrule(lr){7-8}
Model & Method
& Acc. & Pr.
& Acc. & Pr.
& Acc. & Pr. \\
\midrule

\multirow{7}{*}{Qwen3-VL-8B}
& Full Trace
& 74.00 & 0.00
& 42.37 & 0.00
& 60.96 & 0.00 \\
& Random-33
& 70.67 & 34.97
& 38.98 & 46.59
& 57.46 & 34.49 \\
& Random-50
& 60.67 & 46.31
& 44.07 & 54.99
& 58.33 & 50.37 \\
& KL-guided
& 67.33 & 15.66
& 40.68 & 36.08
& 60.96 & 14.15 \\
& Delete All
& 62.00 & 100.00
& 40.68 & 100.00
& 57.02 & 100.00 \\
& Visual Evidence-Only
& 59.33 & 59.47
& 44.07 & 73.69
& 56.58 & 87.40 \\
\rowcolor{spareBg}
& \textbf{SPARE}
& 72.00 & 30.32
& 49.15 & 55.49
& 57.46 & 67.89 \\

\midrule

\multirow{7}{*}{Qwen3-VL-30B-A3B}
& Full Trace
& 67.33 & 0.00
& 45.76 & 0.00
& 36.84 & 0.00 \\
& Random-33
& 61.33 & 28.14
& 52.54 & 35.02
& 17.11 & 34.30 \\
& Random-50
& 68.67 & 46.92
& 61.02 & 51.93
& 16.23 & 50.60 \\
& KL-guided
& 67.33 & 6.93
& 57.63 & 35.69
& 42.11 & 22.97 \\
& Delete All
& 57.33 & 99.42
& 50.85 & 100.00
& 24.56 & 100.00 \\
& Visual Evidence-Only
& 64.00 & 79.94
& 59.32 & 78.84
& 25.44 & 89.68 \\
\rowcolor{spareBg}
& \textbf{SPARE}
& 64.00 & 78.36
& 59.32 & 10.44
& 49.56 & 67.83 \\

\midrule

\multirow{7}{*}{Nemotron-Nano-VL-8B}
& Full Trace
& 45.33 & 0.00
& 40.68 & 0.00
& 26.75 & 0.00 \\
& Random-33
& 44.67 & 32.87
& 38.98 & 6.20
& 16.67 & 30.86 \\
& Random-50
& 48.67 & 64.37
& 38.98 & 36.61
& 16.23 & 49.41 \\
& KL-guided
& 46.00 & 7.36
& 38.98 & 38.07
& 20.61 & 9.06 \\
& Delete All
& 46.67 & 100.00
& 37.29 & 100.00
& 21.49 & 100.00 \\
& Visual Evidence-Only
& 44.00 & 48.53
& 37.29 & 56.08
& 17.98 & 80.14 \\
\rowcolor{spareBg}
& \textbf{SPARE}
& 48.67 & 67.17
& 42.37 & 62.42
& 21.93 & 47.80 \\

\bottomrule
\end{tabular}%
}
\end{table*}

Figure~\ref{fig:accuracy_pruning_tradeoff} compares \textsc{SPARE} with
controls that isolate three aspects of reasoning pruning: whether to prune
indiscriminately, which reasoning steps to prune, and what information to
retain after pruning. \textit{Full Trace} preserves the complete reasoning
history. \textit{Random-33/50} independently removes each completed reasoning
span with probability 0.33 or 0.50, while \textit{Delete All} removes all
historical reasoning. \textit{Visual Evidence-Only} compresses every eligible
reasoning span into structured visual evidence. Finally, \textit{KL-guided}
uses the same summary-conditioned KL signal as \textsc{SPARE}, but removes
only low-KL reasoning and retains high-KL spans in full. All pruning methods
leave the original tool actions, observations, and images unchanged.

\paragraph{Indiscriminate pruning.}
The first group of controls shows that reasoning cannot be removed
indiscriminately. Aggregate accuracy drops from 48.81\% with
\textit{Full Trace} to 41.95\% with \textit{Random-33}, and remains low at
42.64\% with \textit{Random-50} and 42.71\% with \textit{Delete All}.
Although their pruning rates increase from 32.26\% to 50.64\% and 99.93\%,
the three pruning controls perform similarly and substantially below
\textit{Full Trace}. Thus, randomly or uniformly removing reasoning discards
task-relevant information, and the pruning rate alone does not determine
performance.

\paragraph{Which reasoning steps should be pruned?}
Compared with \textit{Delete All}, \textit{KL-guided} improves accuracy from
42.71\% to 48.36\% by retaining high-KL reasoning and removing only low-KL
spans. This confirms the importance of selecting reasoning steps according to
their residual contribution. However, because high-KL spans are retained in
full, \textit{KL-guided} removes only 16.40\% of reasoning, showing that
selection alone is accurate but overly conservative.

\paragraph{What should be retained?}
\textit{Visual Evidence-Only} retains structured visual information instead
of deleting every reasoning span, but improves aggregate accuracy only
marginally over \textit{Delete All} (42.86\% vs.\ 42.71\%). This indicates
that visual-evidence reconstruction alone is insufficient when applied
indiscriminately. In contrast, \textsc{SPARE} applies reconstruction only to
high-KL spans and removes low-KL reasoning entirely. Compared with
\textit{KL-guided}, it increases the pruning rate from 16.40\% to 57.81\%
while improving accuracy from 48.36\% to 50.34\%. These comparisons show that
KL-based selection determines which reasoning steps remain important, while
evidence-preserving reconstruction determines how their useful information
can be retained compactly.

Consequently, \textsc{SPARE} lies above the baseline Pareto frontier in
Figure~\ref{fig:accuracy_pruning_tradeoff}, achieving the highest aggregate
accuracy of 50.34\% while pruning 57.81\% of reasoning tokens. The per-model
results in Table~\ref{tab:complete_accuracy_pruning} further show that this
advantage includes both accuracy-improving and accuracy-preserving cases. On
MNMS with Qwen3-VL-30B-A3B, \textsc{SPARE} achieves 49.56\% accuracy with
67.83\% pruning, compared with 42.11\% accuracy and 22.97\% pruning for
\textit{KL-guided}. On BLINK-Jigsaw with Qwen3-VL-8B, it removes 30.32\% of
reasoning while remaining within two accuracy points of \textit{Full Trace}. These aggregate trends are consistent with the controlled ablations in
Table~\ref{tab:ablation}, which further isolate the complementary roles of
KL-based selection and visual-evidence reconstruction.

\subsection{Computational Cost}
All training and test-time experiments were conducted using eight NVIDIA A100
GPUs. At each pruning event, SPARE generates a task-state summary and performs
two forward replays of the candidate reasoning history to compute the
summary-conditioned KL signal. This introduces temporary inference overhead but
requires neither an auxiliary model nor parameter updates at test time. Adaptive
triggering avoids this cost for short trajectories, while longer trajectories
can amortize it by reusing the reduced context over subsequent interaction
steps.

\section{Supplementary Analyses of Textual Interference}
\label{app:supp_analysis}

\subsection{Controlled Diagnostic of Textual Interference}
\label{app:stale_interference}

\paragraph{Relation to the main experiments.}
The main experiments evaluate the complete \textsc{SPARE} pipeline across
multiple MLLM backbones and full benchmarks. Complementary to those end-to-end
results and the attention analysis, we conduct an additional controlled
diagnostic to test whether pruning stale reasoning makes the model less likely
to follow an obsolete textual hypothesis when later visual or tool evidence
supports a different answer. This diagnostic is separate from the aggregate
benchmark evaluation.

\paragraph{Controlled construction.}
We construct a subset from GTA containing audited cases in which an early
reasoning segment expresses a plausible but incorrect answer and a later tool
observation provides corrective evidence. Across conditions, the original
question, image, tool calls, action blocks, and tool observations are held
fixed; only the retained reasoning history changes.

We compare three conditions. \textbf{Full Trace} keeps the complete history,
including the stale hypothesis. \textbf{Random-1} removes one eligible
reasoning segment at random, providing a control for context shortening without
targeted selection. \textbf{SPARE} automatically applies its pruning
procedure to the eligible reasoning history while preserving the tool
interaction and visual evidence. Every audited conflict case is retained in
the evaluation, including cases in which pruning does not activate and the
context therefore remains unchanged.

\paragraph{Metrics.}
We report forced-choice accuracy (Acc.), which measures how often the model
selects the answer supported by the later evidence, and stale-copy rate (SCR),
which measures how often the final prediction instead repeats the obsolete
textual hypothesis. Formally, for method $m$,
\begin{align}
\mathrm{Acc}_m
&=
\frac{1}{N}
\sum_{i=1}^{N}
\mathbf{1}
\left[
\hat{y}_i^m=y_i
\right],
\\
\mathrm{SCR}_m
&=
\frac{1}{N}
\sum_{i=1}^{N}
\mathbf{1}
\left[
\hat{y}_i^m=s_i
\right],
\end{align}
where $y_i$ and $s_i$ denote the ground-truth and stale answers,
respectively. A lower SCR indicates that the final decision is less likely to
follow obsolete textual reasoning when it conflicts with later evidence.

\begin{table}[h]
\caption{Controlled textual-interference results on the constructed GTA
subset. All values are percentages. Random-1 results are averaged over random
draws; cases without an eligible deletion remain unchanged. SPARE is
evaluated as an automatic procedure rather than with a manually specified
deletion mask.}
\centering
\small
\begin{tabular}{llcc}
\toprule
\textbf{Model} & \textbf{Method}
& \textbf{Acc.} $\uparrow$
& \textbf{SCR} $\downarrow$ \\
\midrule
\multirow{3}{*}{Qwen3-VL-8B}
& Full Trace          & 63.64 & 36.36 \\
& Random-1            & 72.73 & 27.27 \\
& \textsc{SPARE}   & \textbf{90.91} & \textbf{9.09} \\
\midrule
\multirow{3}{*}{Qwen3-VL-30B-A3B}
& Full Trace          & 54.55 & 45.45 \\
& Random-1            & 72.73 & 22.73 \\
& \textsc{SPARE}   & \textbf{81.82} & \textbf{18.18} \\
\bottomrule
\end{tabular}
\label{tab:gta_stale_primary}
\end{table}

\paragraph{Results.}
For Qwen3-VL-8B, SPARE improves forced-choice accuracy from $63.64\%$ to
$90.91\%$ and reduces stale-answer copying from $36.36\%$ to $9.09\%$. For
Qwen3-VL-30B-A3B, accuracy increases from $54.55\%$ to $81.82\%$, while SCR
decreases from $45.45\%$ to $18.18\%$. Thus, across both backbones, SPARE
improves accuracy and reduces stale copying by $27.27$ percentage points.

Random-1 produces smaller improvements than SPARE for both backbones. This
comparison indicates that the benefit cannot be explained solely by shortening
the reasoning history: selecting which reasoning content to remove is important
for reducing interference from stale textual hypotheses. Free-form answer
accuracy remains unchanged between Full Trace and SPARE, suggesting that
the observed effect reflects reduced reliance on conflicting textual history
rather than a general change in task-solving ability.

\paragraph{Scope.}
This deliberately controlled diagnostic is intended as mechanism-level support
for the textual-debt motivation, rather than as a separate benchmark result.
It shows that the automatic SPARE procedure can reduce interference from
stale reasoning while retaining the later visual and tool evidence. Because
the diagnostic isolates a specific form of text--evidence conflict, it should
not be interpreted as a general comparison between SPARE and every possible
reasoning-compression strategy.

\subsection{Attention Visualization After Pruning}
\label{app:visualization}

To further examine whether pruning reduces textual dominance and restores
reliance on visual evidence, we visualize per-layer text-to-image attention
averaged over 10 tool-use trajectories. After applying \textsc{SPARE},
attention to image tokens increases across nearly all layers. This supports our
motivation: redundant accumulated text can dominate the context and distract
the model from visual evidence, while pruning it reallocates computation back
to the image. The cross-layer attention shift provides additional evidence that
suppressing textual dominance restores the role of visual evidence in
downstream reasoning.

\begin{figure}[h]
\centering
\includegraphics[width=0.9\linewidth]{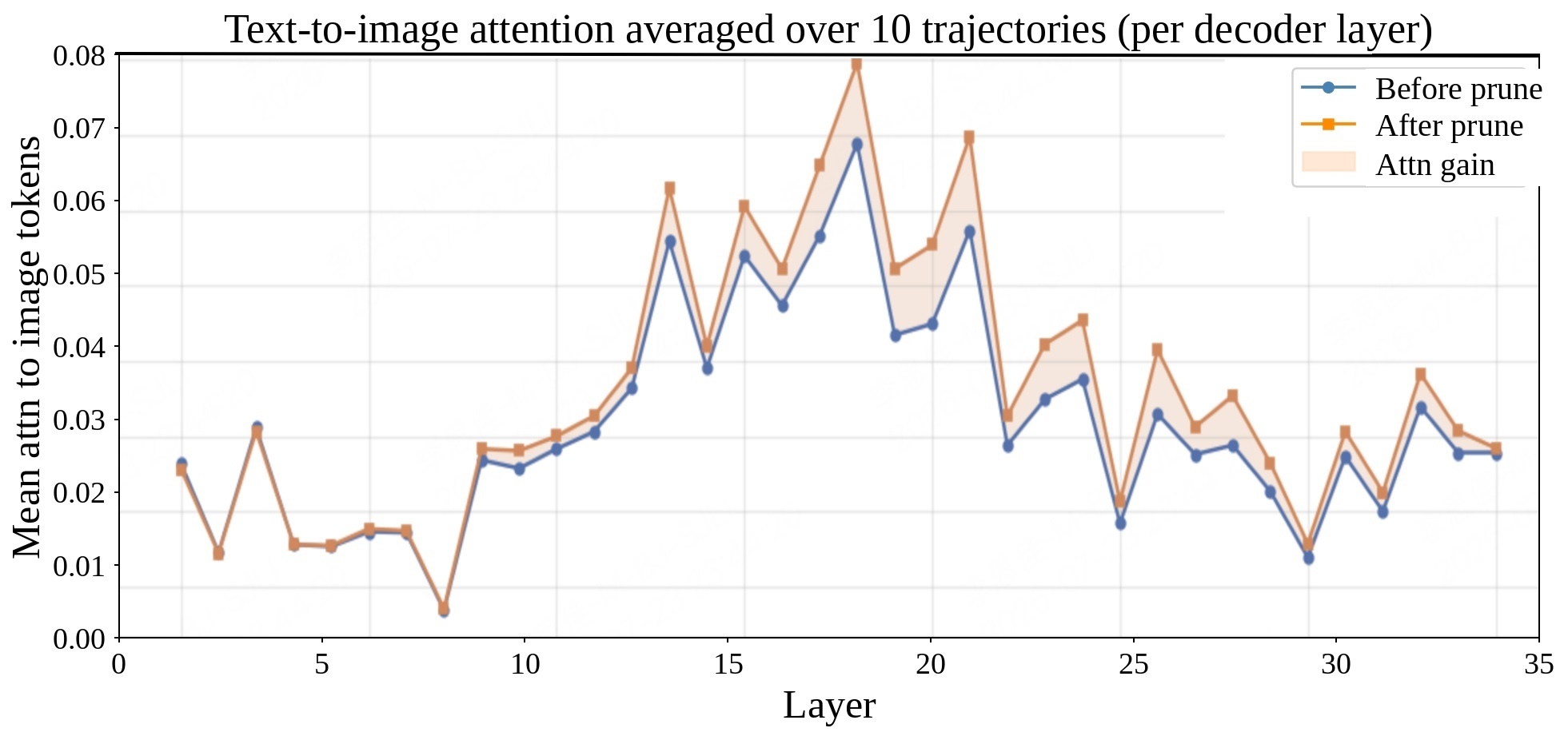}
\caption{Text-to-image attention across decoder layers, averaged over 10 tool-use trajectories. Blue and orange denote attention before and after pruning, respectively, and the shaded region shows the gain. Pruning summary-covered reasoning consistently increases attention to image tokens, indicating stronger reliance on visual evidence.}
\label{fig:attn_motivation_appendix}
\end{figure}

\section{Limitations}
\label{app:limitations}

SPARE is currently designed for multi-step multimodal agents with explicit reasoning and tool-use histories, making its application to single-turn or unstructured agents less direct. It also introduces additional computation for summary generation and context replay; however, adaptive triggering avoids this cost on short trajectories, requires no auxiliary model, and allows the pruned context to be reused in subsequent steps. Although consistent results across three backbones and five benchmarks under the same pruning configuration support its generality, extending SPARE to additional modalities and agent architectures, together with more comprehensive end-to-end latency evaluation, remains an important direction for future work.